\documentclass[a4paper,10pt]{article}
\pdfoutput=1

\usepackage[utf8]{inputenc}
\usepackage{graphicx}
\usepackage{subfig}
\usepackage{textcomp}
\usepackage{listings}
\usepackage{hyperref}

\lstdefinestyle{BashInputStyle}{
  language=bash,
  basicstyle=\small\sffamily,
  numbers=left,
  numberstyle=\tiny,
  numbersep=3pt,
  frame=tb,
  columns=fullflexible,
  linewidth=0.9\linewidth,
  xleftmargin=0.1\linewidth
}

\title{\emph{ira\_laser\_tools}: a ROS LaserScan\\manipulation toolbox}
\author{
    Augusto Luis Ballardini \\
    ballardini@disco.unimib.it
    \and 
    Simone Fontana \\
    simone.fontana@disco.unimib.it\\
    \and 
    Axel Furlan \\
    furlan@disco.unimib.it 
    \and 
    Domenico G. Sorrenti \\
    domenico.sorrenti@unimib.it}
\date{November 04, 2014}

\begin{document}

\maketitle

\begin{abstract}
Laser scanners are sensors of widespread use in robotic applications. Under the Robot Operating System (ROS) the information generated by laser scanners can be conveyed by either LaserScan messages or in the form of PointClouds. Many publicly available algorithms (mapping, localization, navigation, etc.) rely on LaserScan messages, yet a tool for handling multiple lasers, merging their measurements, or to generate generic LaserScan messages from PointClouds, is not available. This report describes two tools, in the form of ROS nodes, which we release as open source under the BSD license, which allow to either merge multiple single-plane laser scans or to generate virtual laser scans from a point cloud. A short tutorial, along with the main advantages and limitations of these tools are presented.
\end{abstract}

\section{Introduction}
The laser tools presented here are designed to ease the handling of multiple laser scanners and their interfacing with those ROS~\cite{288} nodes that require inputs in the form of LaserScan messages. The tools implement the RQT Dynamic Reconfigure interface so to allow an easy and on-line configuration of the node parameters. There are two tools in the ira\_laser\_tools toolbox: the \emph{ira\_laser\_merger} and the \emph{ira\_laser\_virtualizer}.

\emph{ira\_laser\_merger} allows to easily merge multiple single plane laser scans into a single one; this is very useful when using applications like gmapping~\cite{grisettigmapping}, amcl~\cite{ROS-AMCL}, and the navigation stack on vehicles with multiple single plane laser scanners, as these applications require just one laser scan as input. The resulting scan will appear as generated from a single scanner, disregarding the actual occlusions as it would be seen from the merged scan viewpoint. See Section~\ref{merger_limitations}, which is dedicated to this issue.

\emph{ira\_laser\_virtualizer} allows to generate virtual laser scans from a point cloud like the one generated by a multi-plane laser scanner (e.g., a Velodyne laser scanner). The only requirement for this node to work properly is that the roto-translations between the virtual laser scanners and the reference frame of the point cloud must be specified to TF.

Both nodes are provided, along with a commented launch file and a commented dynamic reconfigure file. Please use these launch files as the starting point for your applications.

\section{Installation}

The tools are provided within the ROS package \emph{ira\_laser\_tools} \footnote{\url{https://github.com/iralabdisco/ira_laser_tools}} and are compliant with ROS Indigo Igloo. To install them go to your catkin source directory and clone the repository:

\begin{lstlisting}[style=BashInputStyle]
 cd catkin_ws/src
 git clone https://github.com/iralabdisco/ira_laser_tools
\end{lstlisting}

Then compile the package (make sure you have PCL~\cite{Rusu_ICRA2011_PCL} installed):

\begin{lstlisting}[style=BashInputStyle]
 cd ..
 catkin_make
\end{lstlisting}

That's all. The package is ready. The following sections explain how to use each tool.

\section{\emph{ira\_laser\_merger}}

\subsection{Usage of \emph{ira\_laser\_merger}}
The node has the following parameters, which can be easily modified using the launch files we provide with the nodes:

\begin{itemize}
  \item \textbf{destination\_frame}: the frame the merged scan is referred to.
  \item \textbf{cloud\_destination\_topic}: the topic where the merged scan is published as a point cloud, useful for debugging.
  \item \textbf{scan\_destination\_topic}: the topic where the merged scan is published as a laser message.
  \item \textbf{laserscan\_topics}: the list of the topics to which the node is subscribed.
\end{itemize}

\begin{figure}
  \centering
  \includegraphics[width=0.8\textwidth]{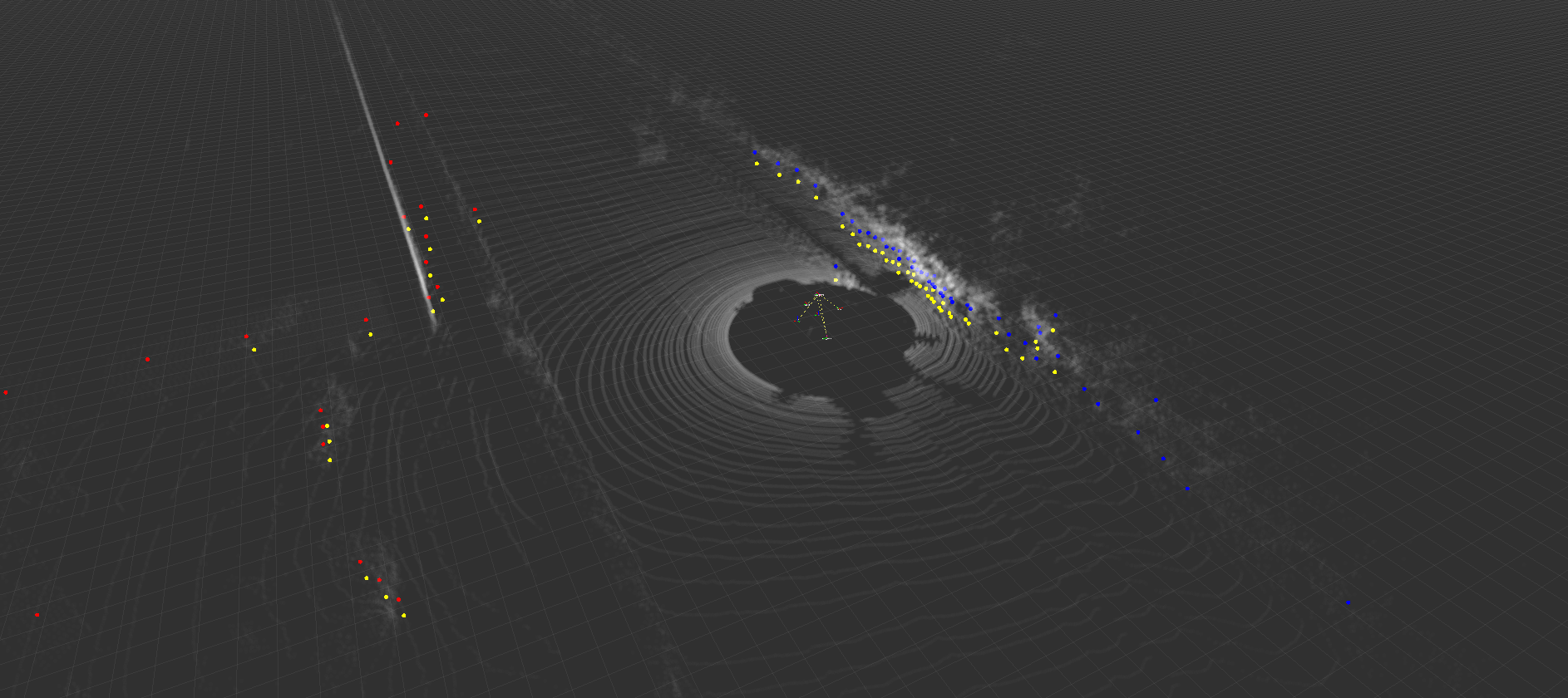}
  \caption{An example of the running laser merger node. Please note how the two input laser scans (red and blue points) are merged in the single output laser scan (yellow). Please notice also that the two input scans are projected on the scanning plane of the output scan, thus resulting in vertical gaps (w.r.t the output scan plane) between the input and output points.}
  \label{fig_merger}
\end{figure}

An example of the node at run-time is shown in Figure~\ref{fig_merger}. Please note how the two input laser scans are merged in the single output laser scan.

\subsection{Limitations of \emph{ira\_laser\_merger}}\label{merger_limitations}

There are two significant limitations to take into consideration, when using this tool.
\begin{itemize}
  \item The generation of the merged laser scan follows an initial phase in which all the laser scans to be merged are flattened to a single plane, which is the scanning plane of the destination (merged) laser scan. As a consequence, objects that would not be observable in the destination (merged) laser scan, but that are in some of the input scans, will (wrongly) appear in the final merged scan.
    \item The output scan must be in the form of the ROS LaserScan message, which implies that the points of the merged scan are generated as if they were measured from the measuring center of the destination laser scanner. Therefore the final merged scan may present data that would not be physically feasible; this might be a problem or not, depending on your application. As an example, consider the case of 2 scanners mounted on the 2 front corners A and B of a rectangular vehicle; each scanner gives out a 270\textdegree{} scan, scanning from backward along the side of the vehicle to laterally toward the other scanner. The merged scan will appear as generated from a virtual scanner positioned in C, halfway A and B, representing, in the merged scan reference frame, the same measures available in the merged scans, irregardless of the vehicle occlusions, which would apply to a real scanner positioned in C.
\end{itemize}

Please be aware of these two limitations when using the \emph{ira\_laser\_merger} tool for your experiments.

\section{\emph{ira\_laser\_virtualizer}}

\subsection{Usage of \emph{ira\_laser\_virtualizer}}
The node has the following parameters, which can be easily modified using the launch files we provide with the nodes:

\begin{itemize}
  \item \textbf{cloud\_topic}: specifies the topic where the original (input) point cloud is published.
  \item \textbf{base\_frame}: the reference frame to which laser(s) scans are referred; it can be the same reference frame as the one of the input cloud.
  \item \textbf{output\_laser\_topic}: if specified, all the virtual lasers are going to be published on this topic; leave empty to publish each virtual laser scan on a separate topic.
  \item \textbf{virtual\_laser\_scan}: list of the desired virtual laser(s) scans. A transform w.r.t. base\_frame must exist in TF for each virtual laser.
\end{itemize}

\begin{figure}
  \centering
  \includegraphics[width=0.8\textwidth]{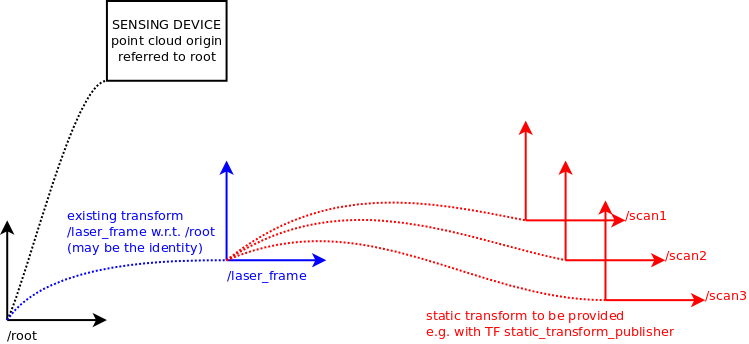}
  \caption{A schematic view of the relationships between the input cloud and the virtual laser scanners.}
  \label{fig_virtualizer_frames}
\end{figure}

In order to create a virtual laser, a transform between the existing point cloud reference frame and the new virtual laser frame has to be provided to the TF node. Please note that this transform has not to be direct, i.e. TF have to be able to retrieve a transform between the two frames, either with a direct transform or using a transform chain. The latter can be extremely useful in configurations like the one shown in Figure~\ref{fig_virtualizer_frames}, where a common reference frame is used to specify several virtual lasers.

\begin{figure}
  \centering
  \subfloat[]{\label{fig:rviz-virtualizer-03} \includegraphics[width=0.8\textwidth]{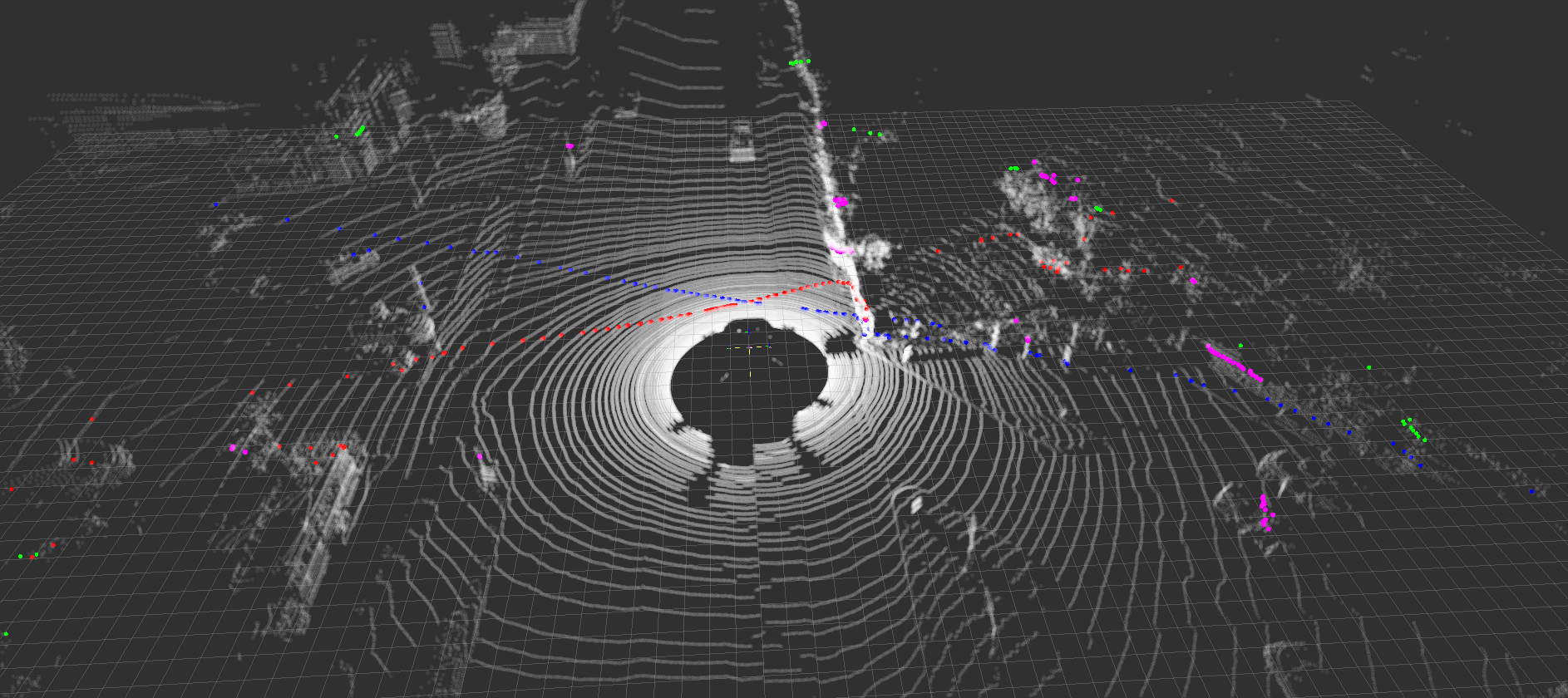}} \\
  \subfloat[]{\label{fig:rviz-virtualizer-02} \includegraphics[width=0.8\textwidth]{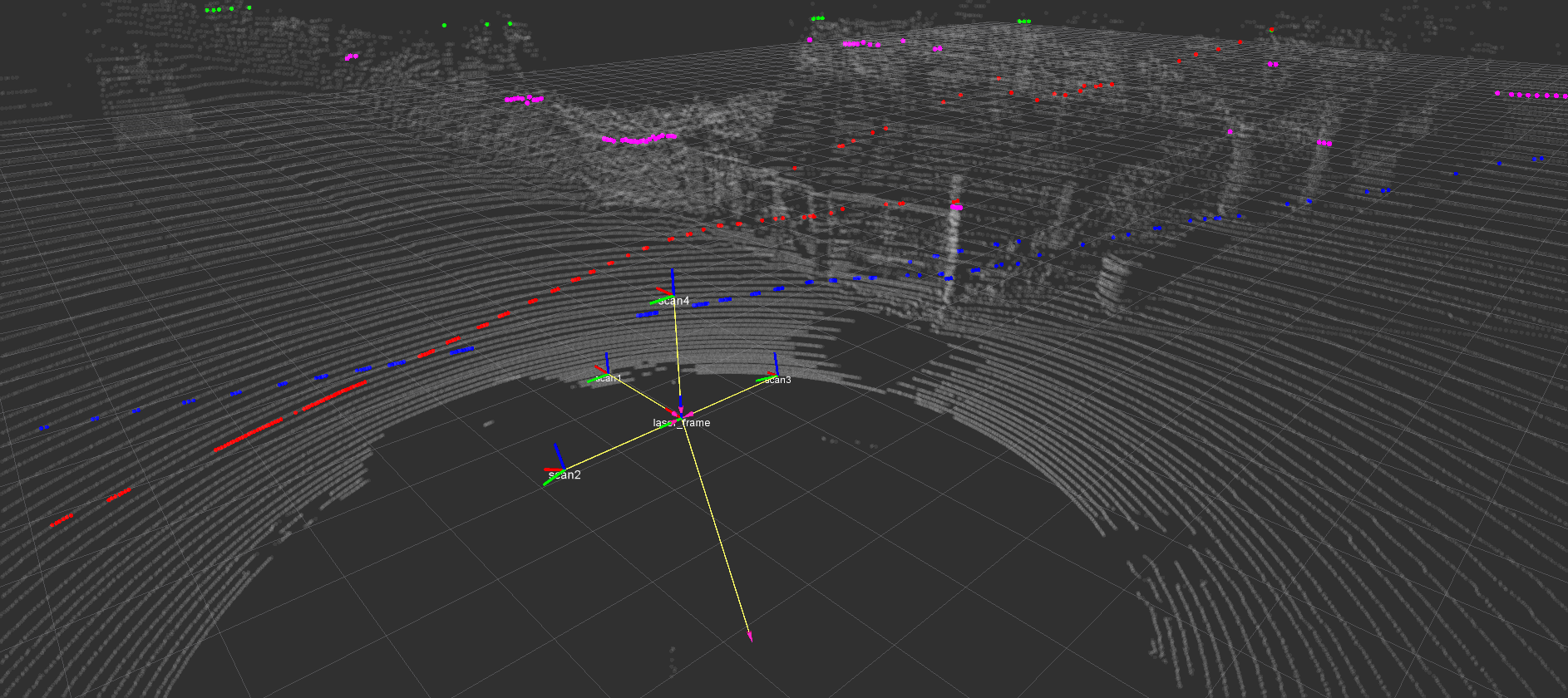}}    
  \caption{An example of the running laser virtualizer node~\protect\subref{fig:rviz-virtualizer-03}. Please note the four virtual laser scans generated from the input point cloud~\protect\subref{fig:rviz-virtualizer-02}}
  \label{fig_virtualizer}
\end{figure}

Since in the most common situations the poses of the virtual laser frames are supposed to be fixed, the ROS Static Transform Publisher can be used, as it easily allows to create static transforms between reference frames. A static transform can be provided using a launch file, specifying the full transforms in terms of translation (X,Y,Z) and rotation (R,P,Y) coordinates, plus a pair of reference frames and a time period for the recurring publishing \footnote{\url{http://wiki.ros.org/tf\#static_transform_publisher}}, as shown in this example:

\begin{lstlisting}
<node pkg="tf"
      type="static_transform_publisher"
      name="ira_static_broadcaster1"
      args="0 0 0 0 0.3 0 laser_frame scan1 1000"
/>
\end{lstlisting}

An example of the node at run-time is shown in Figure~\ref{fig_virtualizer}. Please note the two virtual laser scans generated from the input point cloud.

\subsection{Limitations of \emph{ira\_laser\_virtualizer}}
As in Section~\ref{merger_limitations}, please be aware that the generated virtual laser scans do not take into account possible occlusions and/or artifacts due to large viewpoint differences. As an example, if the point cloud was acquired from one side of the vehicle, a virtual laser scanner placed on the opposite side may not really be able to observe some parts of the scene; conversely, in this virtual case, they would be observed anyway.

\section{Conclusions}
Two tools, in the form of ROS nodes, have been presented, which allow to either merge multiple single-plane laser scans or to generate virtual laser scans from a point cloud. A short tutorial, along with the main advantages and limitations of these tools have been presented as well. The tools are released as open source code under the BSD license.

\section{License}
The code is released under the BSD\footnote{\url{http://opensource.org/licenses/BSD-3-Clause}} license.

\bibliography{references}
\bibliographystyle{unsrt}

\end{document}